# UAV survey coverage path planning of complex regions containing exclusion zones

Shadman Tajwar Shahid[1, *], Shah Md. Ahasan Siddique[1], Md. Mahidul Alam[1]

[1]Department of Mechanical Engineering, Military Institute of Science and Technology, Bangladesh

*Corresponding author: shadmantajwar1997@gmail.com

**Abstract:** This article addresses the challenge of UAV survey coverage path planning for areas that are complex concave polygons, containing exclusion zones or obstacles. While standard drone path planners typically generate coverage paths for simple convex polygons, this study proposes a method to manage more intricate regions that include boundary splits, merges, and interior holes. To achieve this, polygonal decomposition techniques are used to partition the target area into convex sub-regions. The sub-polygons are then merged using a depth-first search algorithm, followed by the generation of wind direction-dependent continuous Boustrophedon paths based on connected components. Polygonal offset by the straight skeleton method was used to ensure a constant safe distance from the exclusion zones. This approach allows UAV path planning in environments with complex geometric constraints.

## INTRODUCTION

Unmanned Aerial Vehicles (UAVs) have become indispensable tools for mapping and surveying due to their ability to provide detailed, high-resolution data over large areas with minimal cost and effort. UAVs are frequently used for remote sensing, capturing geographic and spectral data that can be processed into orthophotos or Digital Elevation Models (DEMs). These outputs offer valuable insights for industries like agriculture, environmental monitoring, and infrastructure management [1].

Image mapping performed by UAVs takes multiple low-altitude aerial images with a gimbaled imaging system. To get high-resolution photos that cover the entire region of interest (ROI), these aerial images are stitched together and geometrically corrected to make highly detailed orthophotos.

The process of generating a path allowing the UAVs to cover every ROI point is called Coverage path planning (CPP). CPP encompasses many applications, including pocket milling, lawn mowing, indoor cleaning robots, and farm vehicle field coverage [2]. In all these cases, the ROI is defined by a polygon. The optimum path for a vehicle to cover a convex polygon is a back-and-forth or zigzag pattern inside the polygon called the Boustrophedon path [3].

In the reviewed literature, it is seen that the authors assume that the ROI is simple convex geometries and overlook the presence of exclusion zones or obstacles. A study conducted in Finland reported that only 13% of agricultural fields are convex. Of those, 25% are simple shapes like circles and rectangles, which means approximately 75% of the field shapes are complex [4]. Therefore, there is a real requirement for the development of versatile CPP algorithms. Hence, this study aims to present a method of automatically generating coverage path planning for UAVs of regions with high geometric complexity.

## METHODOLOGY

The ROI is represented by a 2D polygon consisting of non-intersecting polygonal holes and a polygonal boundary. In path planning, a requirement is that the vehicles follow a continuous path as much as possible, where the vehicle does not have to make unnecessary movements between adjacent paths. So, the vehicles follow a zig-zag pattern across the ROI, covering every area. For a convex polygon i.e. a polygon that has all of its internal angles less than 180 degrees, it is possible to cover the area with a single zig-zag path. However, for a concave polygon, multiple zig-zag paths are needed where the vehicle has to move between each path [5].

## Obstacle/Exclusion zone identification

Obstacles or exclusion zones are called holes in this study and can be explicitly defined as separate polygons within the array or added to a single array containing all line segments. In the latter case, it is necessary to identify and isolate the holes within the polygon. It is assumed that each polygon's line segments are organized in a clockwise or counterclockwise direction, and all polygons form closed chains.

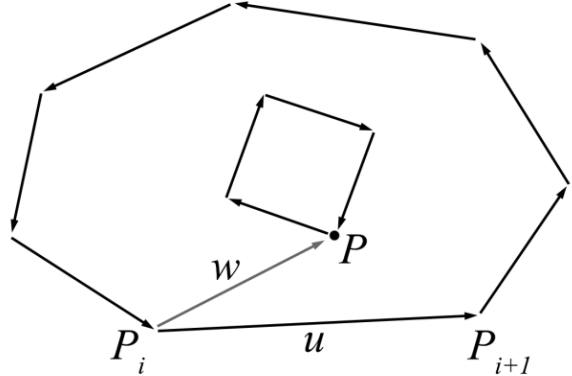

**Figure 1.** Hole identification by cross-product.

$$u = P_{i+1} - P_i, w = P - P_i \tag{1}$$

Holes within a polygonal chain are identified by testing the orientation of all vertices in one chain against the line segments of other chains. The sidedness of the points is determined using the cross-product, $u \times w$. If all points of one chain lie inside another chain, it is designated as a hole corresponding to the outer chain. If the hole belongs to an odd number of outer chains, it is an inner boundary. In this case, the direction of the inner polygon is reversed to that of the outer polygon because they are offset outwards, whereas the outer polygon is offset inwards.

## Polygonal offset

The obstacle or exclusion zone in the polygon needs to be offset outwards by a safe distance because following exactly the boundary of the exclusion zone would cause collision or trespassing into the exclusion zone. Moreover, according to requirements, the ROI boundary or the outer polygon may need to be offset inwards to maintain a constant safe distance.

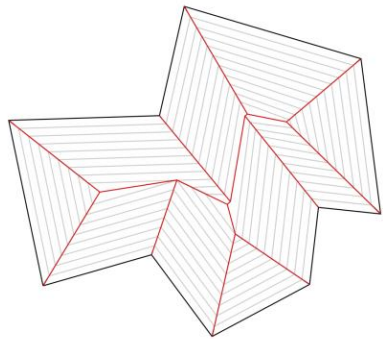

**Figure 2.** Offsets are produced by the straight skeleton method. In red, are the edge-front interfaces defining the straight skeleton.

The offset of the polygon is done by the straight skeleton (SS) method which produces offsets by wavefront propagation [6, 7]. The SS generates new polygons by tracing the outward or inward movement of the polygon's edge moving at unit speed, parallel to the edge. The edges may propagate until they meet and form vertices. Points along the edge-front, equidistant from the polygon are connected to produce the offset. This method inherently avoids creating intersecting geometries, which is especially useful when the polygon consists of dense geometries and concave angles. Using the SS method to offset polygons maintains sharp corners and preserves the overall shape of the polygon.

## Polygonal decomposition

The boundary for the zig-zag paths is the polygon that is not a hole of any other polygon. The polygon is tested for concavity by checking if any of the internal angles is more than 180 degrees. If all angles are less than 180 degrees, and no holes correspond to the polygon, then the polygon is convex, and zig-zag paths can be generated directly. Otherwise, polygonal decomposition is necessary.

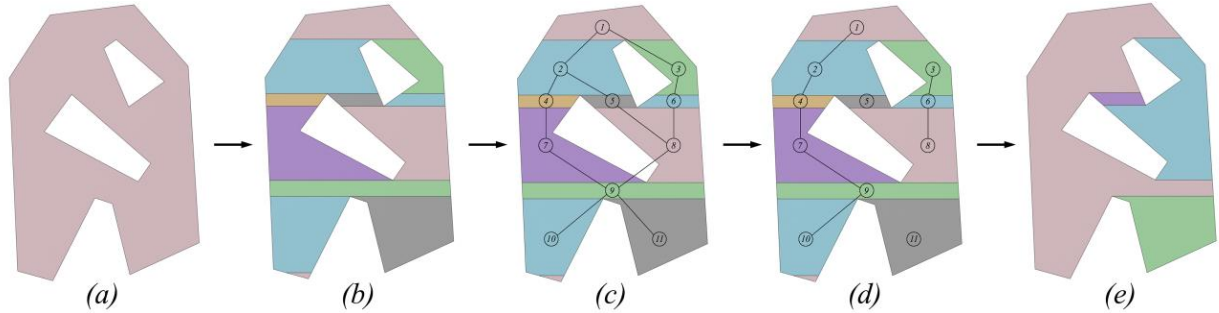

**Figure 3.** a) Boundary to generate zigzag infill b) Partitioning polygon by sweep line algorithm c) Creation of adjacency graph d) Creation of different connected components from adjacency graph e) Connecting polygons based on connected components.

In a concave polygon, a merge event occurs when two edges converge inwardly at a vertex, and a split event occurs when two polygon edges diverge from a vertex. The merge and split events are identified using the sweep line method [8]. The sweep line's angle is crucial for optimizing the efficiency of the UAV's coverage path. The slope or angle of the sweep line relative to the longitude and latitude of the earth depends on the wind speed and direction. This is because a UAV will have faster speed in each portion of the boustrophedon path when the sweep angle is perpendicular to the wind, rather than a sweep angle parallel to the wind direction [9].

Firstly, the line segments of the polygons are sorted based on their decreasing y-coordinates. For a zigzag pattern that is not parallel to the x-axis, which is usually the case, the polygon needs to be transformed with the corresponding rotation matrix. Transforming the original polygon before polygonal decomposition reduces the number of computations since the slope of the sweep line is always 0. In this case, the resultant path must be transformed back to the original axis. A horizontal line segment is swept from the maximum to the minimum y-coordinate, with the number of intersections between the line and the polygon edges being counted. The number of intersections above and below each vertex is compared at each point and the points where the number of intersections changes correspond to merge or split events. The lines passing through these event points partition the polygon into convex sub-polygons with respect to the sweep line.

Once the concave polygon is decomposed, each convex sub-polygon is treated as a separate entity, for which individual zig-zag (boustrophedon) paths can be generated. Longer paths can be created by merging the sub-polygons. This is done by making an adjacency graph. In this graph, each node represents a sub-polygon, and an edge exists between two nodes if their corresponding sub-polygons share a partition edge. Shared edges are identified based on common vertexes. The next step is to identify connected components in the adjacency graph. A depth-first search (DFS) algorithm generates the connected components. Starting from any unvisited node (sub-polygon), the DFS visits all nodes (adjacent sub-polygons) to form a connected component. The process is repeated until all sub-polygons have been visited and added to one of the connected components. Finally, the sub-polygons are merged in the order the connected components give [10].

## Zig-zag path generation

The connected polygons are the boundaries for generating individual zig-zag infill patterns. The slope of these paths must be the same as that of the sweep line. Since the connected polygons are convex, each sweep line intersects the polygon at two points, and these points are connected to the previous line's points in an alternating direction to form the zig-zag paths. The equations for the parallel lines are as follows:

$$y = mx + n \times d\sqrt{m^2 + 1} \qquad (2)$$

Where d is the distance between each path, m is the slope of the lines, and n is the sequence number of the line. To ensure optimal image overlap, the distance d must be chosen such that there is an 80% overlap between adjacent images. The final output is a series of longer boustrophedon paths that cover the entire area of the original concave polygon. The process of generating Boustrophedon paths is discussed in algorithm 1.

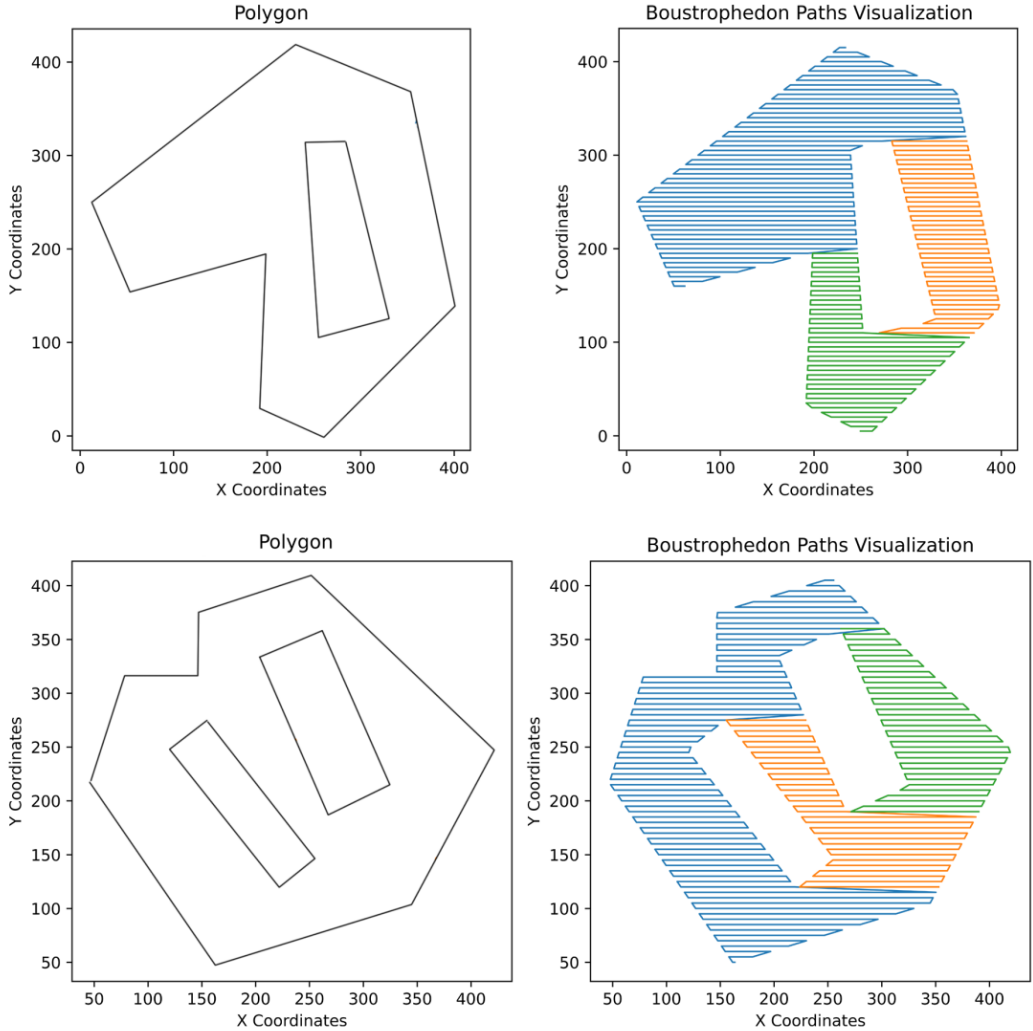

**Figure 4.** Examples of polygonal decomposition of concave polygon. On the left is the original polygon and on the right are Boustrophedon paths produced after polygonal decomposition.

**Algorithm 1:** Generation of Boustrophedon paths from the region of interest

**Step 1:** Polygon decomposition

**Input:** ROI boundary: *boundary*

**Output:** Connected sub-polygons: *polygons*

Find the boundary of the region by offsetting *boundary* by offset distance *d* to *boundary* using the straight skeleton method

Sort *boundary* in terms of y-coordinate values

For *i* from 0 to length(*boundary*):

Find the number of intersections of the line along vertices of *boundary* with edges of *boundary* and add to *intersection_number*

For i from 1 to length(intersection_number) – 1:

    If *intersection_number[i]* != *intersection_number[i-1]* or *intersection_number[i]* != *intersection_number[i+1]*:

        Add y-coordinate in *boundary [i]* to *partition_edges*

For *i* from 1 to length(*partition_edges*):

    Find vertices above and on *partition_edges[i]* and add to *sub_polygons*

    Delete all vertices above *partition_edges[i]* in *the boundary*

Initialize an empty graph *G*.

For *i* from 0 to length(*sub_polygons*):

    For *j* from *i*+1 to length(*sub_polygons*):

        Find the number of shared vertices between *sub_polygons[i]* and *sub_polygons[j]* and add to *shared_vertices*

        If *shared_vertices* >= 2:

            Add an edge between *i* and *j* in graph *G*

Initialize empty array *paths*, *connected_components*, and *visited*

For *i* from 0 to length(*sub_polygons*):

    If *visited[i]* == FALSE:

        Add *i* to *connected_ components*

        *visited[i]* == TRUE

        For *j* from 0 to length(*sub_polygons*):

            If *visited[j]* == FALSE and an edge exists between *i* and *j* in graph *G*:

                Add *j* to *connected_ components*

                *visited[j]* = TRUE

        Add *connected_ components* to *paths*

For *i* from 0 to length(*paths*):

    For *j* from 0 to length(*paths[i]*):

        Add *sub_polygons[path[i][j]]* to *polygons*

**Step 2:** Generation of Boustrophedon paths from connected components

**Input:** Connected sub-polygons: *polygons*

**Output:** Boustrophedon paths: *boustrophedon_paths*

Delete shared edges in *polygons*

For *i* in *polygons*:

    Find minimum and maximum y co-ordinates in *polygons[i]*, *y_min* and *y_max*

    Initialize *y = y_min*

    Initialize *n = 1*

If *y* >= *y_min* and *y* <= *y_max*:

    Find the intersection of line *y* = *y_min* + (*path_spacing* * *n*) with *polygons[i]*

    If *n* % 2 != 0:

        Add intersection points to *boustrophedon_paths*

    Else:

        Swap and add intersection points to *boustrophedon_paths*

    *y* = *y_min* + (*path_spacing* * *n*)

    *n* += 1

## RESULTS AND DISCUSSION

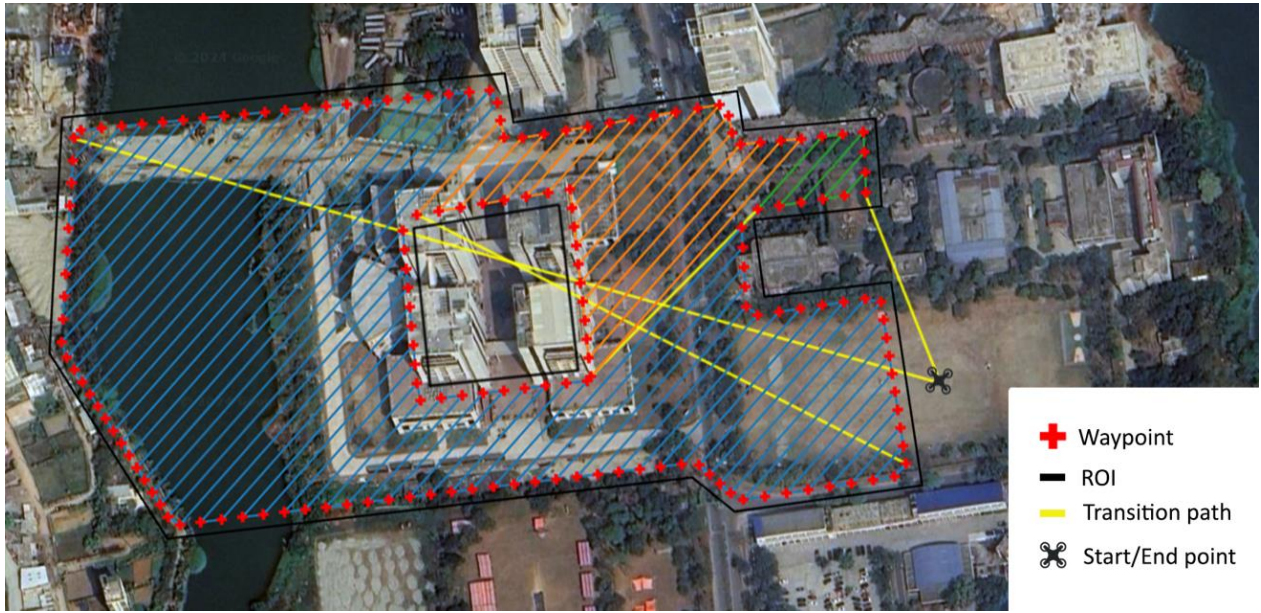

**Figure 5.** Flight path overlaid on satellite image.

The proposed method for UAV survey coverage path planning was tested on complex regions containing concave angles and exclusion zones. As demonstrated in Figure 5, the algorithm successfully generated boustrophedon paths for a real-world region of interest (ROI) while avoiding collisions with exclusion zones. With a wind direction at 45 degrees relative to the reference axis, the path planning process required the generation of three individual boustrophedon paths, with transition paths connecting them. Achieving this using standard methods would have required manually defining three separate convex polygonal regions, which is difficult when accounting for wind direction. Moreover, the example in Figure 4 shows the algorithm's ability to handle multiple exclusion zones. The polygonal offset, implemented through the straight skeleton method, maintained a constant safe distance from obstacles, while the zigzag paths minimized unnecessary UAV turning and repositioning.

## CONCLUSIONS

The proposed UAV survey coverage path planning method effectively addresses the challenges of navigating complex regions containing exclusion zones. By using polygonal decomposition and the polygonal offset, the algorithm generated collision-free boustrophedon paths while maintaining a constant safe distance from obstacles. Unlike standard methods, which would have required manual instructions to define separate regions, the proposed method automated this process. Overall, this solution allows efficient UAV surveys in complex environments, making it practical and streamlined for real-world applications.